\pdfoutput=1

\documentclass[11pt]{article}

\usepackage{EACL2023}

\usepackage{times}
\usepackage{latexsym}
\usepackage{graphicx}
\usepackage{lipsum}
\usepackage{url}
\usepackage{amsmath}
\usepackage{latexsym}
\usepackage{multirow}
\usepackage{rotating}
\usepackage{tabularx}
\usepackage{supertabular}
\usepackage{subcaption}
\usepackage{graphicx}
\usepackage{caption}
\usepackage{subcaption}
\usepackage{verbatim}
\usepackage{tabularx}
\usepackage{supertabular}
\usepackage{amssymb,amsmath,array}
\usepackage{multirow}
\usepackage{rotating}
\usepackage{subcaption}
\usepackage{graphicx}
\usepackage{caption}
\usepackage{subcaption}
\usepackage{lscape}

\usepackage[T1]{fontenc}

\usepackage[utf8]{inputenc}

\usepackage{microtype}

\usepackage{inconsolata}

\newcommand\footnoteref[1]{\protected@xdef\@thefnmark{\ref{#1}}\@footnotemark}

\title{Empowering Cross-lingual Abilities of Instruction-tuned Large Language Models by Translation-following demonstrations}

\author{\textbf{Leonardo Ranaldi $^{(\dagger)}$,  Giulia Pucci, André Freitas$^{(\dagger,*)}$} \\
	${(\dagger)}$ Idiap Research Institute, Martigny, Switzerland \\
 ${(*)}$Department of Computer Science, University of Manchester, UK \\
		{
  {\tt [first\_name].[last\_name]@idiap.ch},
  {\tt [first\_name].[last\_name]@manchester.ac.uk}
  } } 

\begin{document}
\maketitle
\begin{abstract}
The language ability of Large Language Models (LLMs) is often unbalanced towards English because of the imbalance in the distribution of the pre-training data. This disparity is demanded in further fine-tuning and affecting the cross-lingual abilities of LLMs. In this paper, we propose to empower Instruction-tuned LLMs (It-LLMs) in languages other than English by building semantic alignment between them. Hence, we propose \textit{CrossAlpaca}, an It-LLM with cross-lingual Instruction-following and Translation-following demonstrations to improve semantic alignment between languages. We validate our approach on the multilingual Question Answering (QA) benchmarks XQUAD and MLQA and adapted versions of MMLU and BBH. 
Our models, tested over six different languages, outperform the It-LLMs tuned on monolingual data. The final results show that instruction tuning on non-English data is not enough and that semantic alignment can be further improved by Translation-following demonstrations. Our code and data are available \footnote{\url{https://github.com/lranaldii/CrossAlpaca}} 
\end{abstract}

\section{Introduction}

Large Language Models (LLMs) achieve comprehensive language abilities through pre-training on large corpora \cite{brown2020language,touvron2023llama}. 
Hence, the acquired language abilities follow from the corpora features, which are mostly available in English \cite{lin2021pretraining,zhang2023bayling,zhu2023multilingual}. This phenomenon produces an imbalance in both pre-training \cite{Blevins2022LanguageCH} and fine-tuning \cite{le2021lightweight}. Consequently, LLMs underperform in languages different from English \cite{huang2023languages,bang2023multitask}. 

Efforts to increase multilingual abilities propose continuing pre-training with large-scale monolingual data \cite{imani2023glot500,cui2023efficient,yang2023bigtranslate}. However, learning a language from monolingual data requires considerable data and computational resources.

In this paper, we propose \textit{CrossAlpaca}, an Instruction-tuned LLM (It-LLM) empowered with a semantic alignment between English and other languages. We analyze how to elicit the non-English abilities of It-LLMs by focusing on the crucial phase: instruction-tuning over Instruction-following demonstrations. Hence, we study the effect of cross-lingual alignment by proposing Translation-following demonstrations to improve the instruction-tuning phase.

\begin{figure*}[t!]
\centering
    \includegraphics[width=1.1\textwidth]{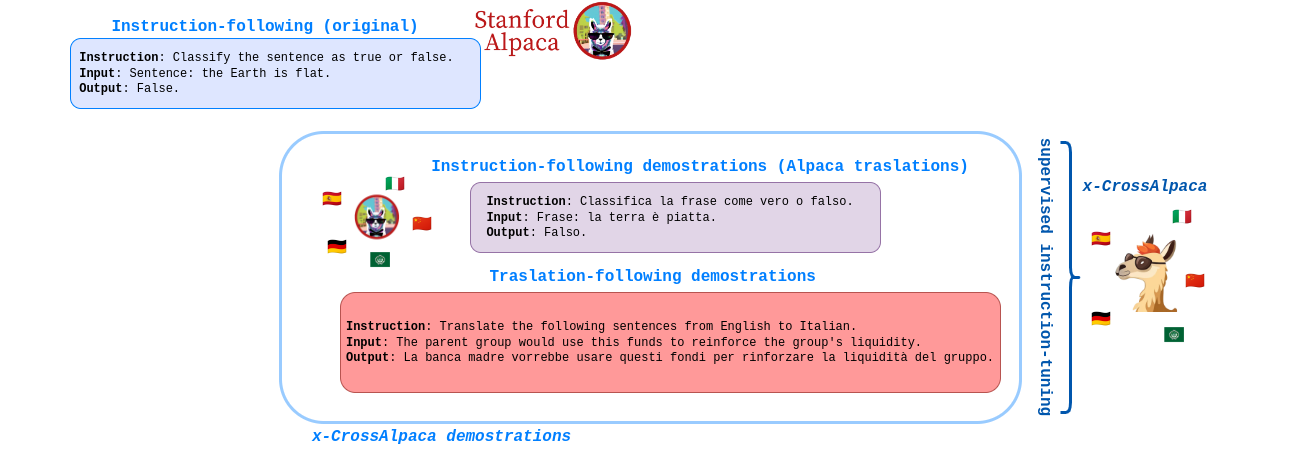}
    \caption{Our \textit{x-CrossAlpacas} are fine-tuned on Instruction-following and Translation-following demonstrations. This example shows the \textit{it-CrossAlpaca}, fine-tuned on it-Alpaca and Translation-following demonstrations.}
    \label{fig:our_proposal}
\end{figure*}

In our experiments, we use \textit{LLaMA-7B} \cite{touvron2023llama} as the LLM backbone and consider six target languages selected from among the available data (shown in Table \ref{tab:details_data_available}); where data are absent, we perform a translation task. As Instruction-following demonstrations, we use the Stanford Alpaca dataset \cite{alpaca} and translated versions in the corresponding languages, while for the Translation-following, we use a publicly available translation resource news\_commentary \cite{tiedemann-2012-parallel}, the most accessible and extendable to multiple languages (i.e., \textit{CrossAlpaca demonstrations} on Figure \ref{fig:our_proposal}). 

Behind an instruction-tuning phase, we evaluate the performance of our six language-specific \textit{CrossAlpacas} using four different benchmarks: two native-multilingual, i.e., XQUAD \cite{Artetxe2019OnTC} and MLQA \cite{lewis-etal-2020-mlqa}, and two native-monolinguals, MMLU \cite{hendrycks2021measuring} and BBH \cite{suzgun2022challenging}. The results show that \textit{CrossAlpacas} instructed with language-specific instruction and translation data far outperform Alpacas instructed only with non-English demonstrations. However, even though \textit{CrossAlpaca} reduces the gap, the Translation-following models of the original Alpaca still perform best. This result shows that LLaMA's learning abilities on English data are superior to those on non-English data. The semantic alignment task also improves the cross-lingual abilities of It-LLMs.

Our findings can be summarized as follows: 
\begin{itemize}
    \item The learning abilities of LLMs on non-English Instruction-tuning tasks are limited;
    \item The multi-lingual abilities of Instruction-tuned LLMs could be empowered through cross-lingual alignment;
    \item Thus, we propose \textit{CrossAlpaca}, an Instruction-tuning approach for non-English models that are based on Instruction-following and Translation-following demonstration for target language; 
    \item  We show that It-LLMs are able to semantically align through cross-lingual Translation-following demonstrations via an extensive evaluation on four multi-lingual QA benchmarks.
\end{itemize}

\section{Related Work}

\subsection{Pre-trained Language Model on Cross-Lingual Corpora}

The next token prediction based on the prefix sequence, as well-known as language modeling, is the everlasting task of modern NLP \cite{tenney-etal-2019-bert}. The extensive knowledge of today's Large Language Models (LLMs) depends on the billions of neurons trained on large-scale corpora with derivatives of the language modeling task. Consequently, the pre-training corpora are predominantly in English, e.g., BooksCorpus \cite{Zhu_2015_ICCV}, MEGATRON-LM \cite{Shoeybi2019MegatronLMTM}, Gutenberg Dataset
 \cite{lahiri:2014:SRW} PILE \cite{gao2020pile}, C4 \cite{dodge2021documenting}, RefinedWeb \cite{penedo2023refinedweb}; therefore, LLMs usually have much better knowledge of English than other languages.

\citet{aulamo-tiedemann-2019-opus,abadji-etal-2022-towards}, in order to solve this problem, propose forward corpora translated into several languages.
However, these corpora are not as huge as their competitors, and the lack of massively parallel data in the pre-training corpora also prevents LLMs from aligning the different languages well \cite{li2023eliciting}. 

\begin{figure*}
\centering
    \includegraphics[width=.75\textwidth]{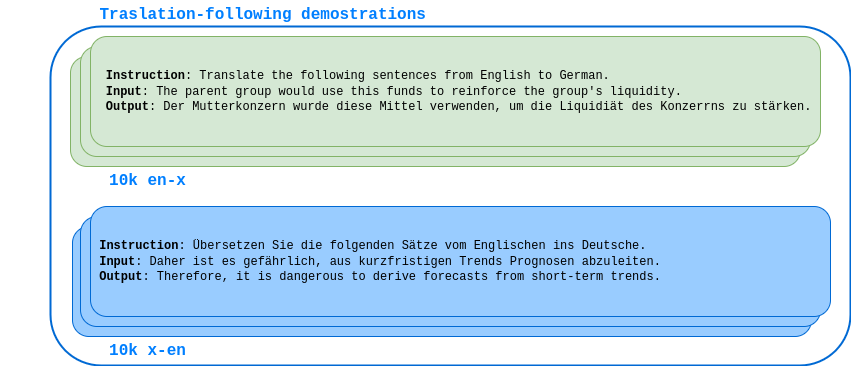}

 \caption{Examples of Translation-following demonstrations. In this particular example, there are two demonstrations with two different directions from English to German (en-x) and vice versa.}
 \label{fig:traslation_following_demo}
\end{figure*}

\subsection{The Instruction-tuning Paradigm}
\label{sec:Instruction-tuning-Paradigm}
\citet{ouyang2022training,wei2022finetuned} fine-tuned LLMs using the Instruction-tuning method based on Instruction-tuning data, which are instruction-response corpora, to make LLMs more scalable and improve zero-shot performance. In this method, the LLM backbone is fed with data from the instruction ${(I,X,Y)}$, where $I$ is an instruction describing the task's requirements, $X$ is the input, which can be optional, and $Y$ is the output for the given task. The goal of this method is to minimize the function $f(Y)$: 
\begin{equation}
   f(Y) =  \arg\min_{\theta} \log p_{\theta}(Y \mid I, X)
\end{equation}
where $\theta$ are model learnable parameters.

Earlier studies show that the instruction-tuning method of LLMs with both human \cite{wang2023selfinstruct} and synthetic-generated instructions \cite{alpaca,xu2023baize} empowers the ability of LLMs to solve considerable tasks in zero-shot scenarios.

However, we state that the generally used instruction-tuning datasets, alpaca \cite{alpaca}, Self-Instruct \cite{wang2023selfinstruct}, Self-Chat \cite{xu2023baize}, conceived in English, which limits the prospect of LLMs to follow non-English instructions and therefore solve related tasks.

\subsection{Instruction-tuning is at hand}
\label{sec:optimization}
Large Language Models have considerable success with many techniques in vogue, such as Instruction-tuning. However, their prohibitive size only allows part of the scientific community to experiment with these models. 

The latest advances that make these models and techniques accessible involve efficient parameter tuning. Parameter-Efficient Tuning (PEFT) is an efficient technique to adjust a small part of the model parameters and freeze the others.   
The main goal is to significantly reduce computational and storage costs while maintaining the performance of the original models. The standard techniques for PEFT are LoRA \cite{hu2022lora}, Prefix Tuning \cite{li-liang-2021-prefix}, P-Tuning \cite{liu-etal-2022-p}. The basic idea is to keep the pre-trained model weights and incorporate low-rank matrices in each architecture layer. This approach significantly reduces the number of parameters that require training for subsequent tasks, thereby increasing overall efficiency. This building block has been and continues to be very important because it skills the scientific community to fair research and enables the development of multiple open-source works.

\subsection{Cross-lingual Instruction-tuning Challenge}
Recent work has shown the remarkable abilities of LLMs in learning instruction in different languages.
\citet{santilli2023camoscio,traditional-chinese-alpaca} proposed monolingual fine-tuning LLaMA adaptations on language-specific translated instructions. The use of optimization techniques introduces in Section \ref{sec:optimization}, to propose engaging custom adapters for various tasks. \citet{zhang2023bayling,zhu2023extrapolating} investigated the cross-lingual abilities of It-LLMs by promoting the effects of augmenting instruction demonstrations. In particular, \citet{zhang2023bayling} use fine-grained custom translations that are not easily accessible, while \citet{zhu2023extrapolating} propose a multilingual approach by considerably increasing instruction and translation tasks. However, these studies have opened an interesting avenue in analyzing the cross-lingual abilities of It-LLMs. 

In this paper, we take the next step by proposing our \textit{CrossAlpaca}. Our methods show of the cross-lingual learning abilities of It-LLMs on a large scale.  Using four well-known benchmarks, we show that the weaknesses of It-LLMs trained on non-English data can be strengthened with cross-lingual alignment approaches. Hence, our analyses aim to understand the role of Instruction-following and Translation-following demonstrations in closing the gap in learning and adapting LLMs' abilities to non-English languages.

\begin{table}
\tiny
\centering 
 \begin{tabular}{l|ll}
\textbf{Model}  & \textbf{Language} & \textbf{Name} \\

\hline
Alpaca \cite{alpaca}  & English &  en-Alpaca \\

Alpaca-Chinese \cite{traditional-chinese-alpaca}  & Chinese & zh-Alpaca  \\

Arabic Alpaca \cite{yasbok-alpaca-instruction-fine-tune-arabic} & Arabic & ar-Alpaca  \\ 

Camoscio \cite{santilli2023camoscio}  & Italian & it-Alpaca  \\

Guanaco \cite{Guanaco-lora} & Spanish & es-Alpaca  \\

German Alpaca \cite{de_alpaca} & German & de-Alpaca  \\

 \end{tabular}

 \caption{Details and names of mono-lingual Instruction-tuned Large Language Models that use a language-specific version of Alpaca as instruction-tuning data.}
 \label{tab:details_mono_ItLLMs}
\end{table}

\section{Data \& Methods}

Pre-training from scratch a Large Language Model (LLM) in multiple languages is cost-prohibitive for data collection and parameter learning. This is the reason why the trend is to do further fine-tuning to strengthen the models' abilities in a specific language \cite{tanti2021languagespecificity,moslem2023adaptive}. In this paper, we aim to exploit the abilities of pre-trained LLMs for non-English languages by further improving the alignment between English and the target language. Hence, in Section \ref{sec:Mono-It-LLMs}, we introduce the challenge of fine-tuning an LLM on a mono-lingual custom-scenario. Furthermore, in Section \ref{sec:Cross-It-LLMs}, we propose our approach to fine-tune cross-lingual LLMs.

\subsection{Mono-lingual Instruction-tuning}
\label{sec:Mono-It-LLMs}
The limited accessibility and transparency of paid API services of state-of-the-art LLMs have pushed research toward developing open-source models. Using the instruction-tuning paradigm (Section \ref{sec:Instruction-tuning-Paradigm}) and Stanford Alpaca \cite{alpaca}, a corpus consisting of 52k of English instruction-output pairs generated by \texttt{text-davinci-003}, several Instruction-tuned versions of LLaMA were released. 

Following this approach, multiple mono-lingual Instruction-tuned versions of LLaMA were proposed by translating the Stanford Alpaca Instruction-following data into the specific language. 
Table \ref{tab:details_data_available} (Appendix \ref{sec:Appendix3}) shows a set of alpaca versions available as open source. Following a systematic analysis of the translated versions of Alpaca in official repositories\footnote{official versions on \url{https://github.com/tloen/alpaca-lora} and \url{https://huggingface.co/models}}, the languages of the benchmark datasets, and the translation pairs present in news\_commentary, which will be introduced later, we selected the languages that share the most already available data. Table \ref{tab:details_mono_ItLLMs} shows the custom versions used in this work, which for simplicity will be renamed \textit{x}-Alpaca, where \textit{x} indicates the specific language.

\begin{figure*}[ht]
\centering
         \begin{minipage}{0.45\linewidth}
     \centering
     \includegraphics[width=\linewidth]{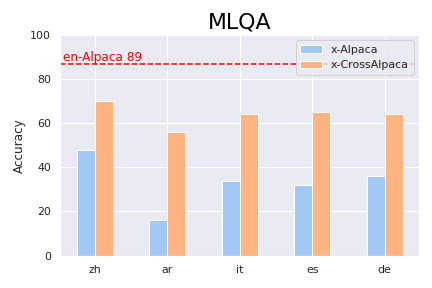}
   \end{minipage}
            \begin{minipage}{0.45\linewidth}
     \centering
     \includegraphics[width=\linewidth]{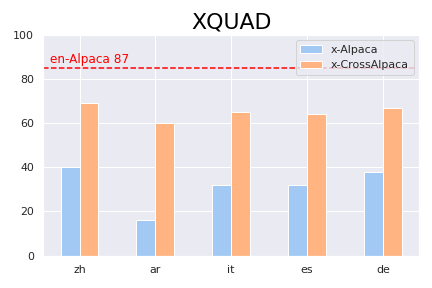}
   \end{minipage}
         \begin{minipage}{0.45\linewidth}
     \centering
     \includegraphics[width=\linewidth]{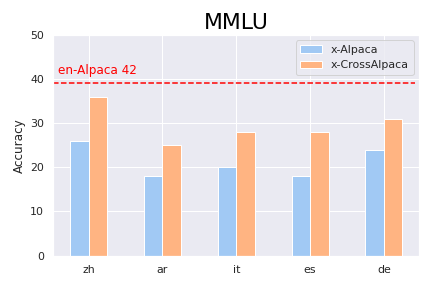}
   \end{minipage}
            \begin{minipage}{0.45\linewidth}
     \centering
     \includegraphics[width=\linewidth]{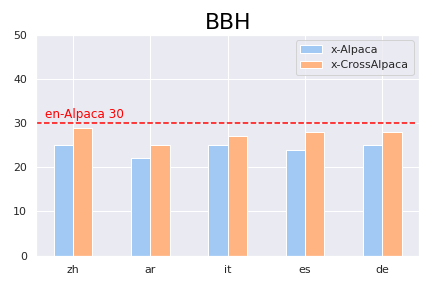}
   \end{minipage}
   \caption{Evaluation results on proposed benchmarks. The dotted line represents the performance of the original Alpaca \cite{alpaca} on English tasks.} 
   \label{fig:performances}

\end{figure*}

\subsection{Cross-lingual Instruction-tuning}
\label{sec:Cross-It-LLMs}

Mono-lingual Instruction-tuning approaches (Section \ref{sec:Mono-It-LLMs}) reward LLMs' multi-lingual abilities. However, the use of translated Alpaca alone for the specific language is not sufficient to elicit the non-English ability of LLMs. For this reason, we propose \textit{CrossAlpaca} (Figure \ref{fig:our_proposal}), which enriches the cross-lingual Instruction-tuning with the Translation-following demonstrations. We aim to highlight LLMs' English and non-English abilities by proposing a semantic alignment task to enrich versions of Alpaca. 


\paragraph{Instruction-following demonstrations} 
The original version of Alpaca is in English. However, as described in Section \ref{sec:Mono-It-LLMs}, different open-source translations are produced with a translation engine. In our experiments, we propose the Instruction-tuning phase with the original English version (en-Alpaca) and the translated language-specific versions. Moreover, we propose the \textit{CrossAlpaca} versions built with the different Alpacas translated into the specific language and the cross-lingual translations (introduced later). In this way, we investigate the abilities of the LLM backbone in understanding multilingual instructions and cross-lingual alignment.

\paragraph{Translation-following demonstrations} 
Learning general instruction data is a sensible strategy for building models to solve multi-tasks directed by instructions \cite{wang2023selfinstruct,zeng2023tim}. However, translation data could contribute to learning semantic alignment. \citet{zhang2023bayling} showed that the translation performance of LLMs can be improved by using expertly annotated translation data, and \citet{zhu2023extrapolating} showed that it could be beneficial for instruction-tuning. 

Inspired by \citet{zhu2023extrapolating}, we use publicly available sentence-level translation datasets, such as \textit{news\_commentary} \cite{tiedemann-2012-parallel}, to construct the translation task instruction demonstrations. We also propose extending this to additional languages, which we release as an open-source dataset. In particular, for each specific language, we constructed 20k demonstrations. Hence, following the Alpaca style (Instruction, Input, and Output) (see Figure \ref{fig:our_proposal}), we randomly selected 10k English to non-English translations and 10k random non-English to English translations (all translations come from news\_commentary).
We have constructed these correspondences for the languages mentioned above in Section \ref{sec:Mono-It-LLMs}. Figure \ref{fig:traslation_following_demo} shows a case of Translation-following for English-German direction (en-de) and vice versa (de-en).

\section{Experiments}
In order to observe the English and non-English abilities of Large Language Models (LLMs) and the impact of the Instruction-tuning approach in cross-lingual scenarios, we propose \textit{CrossAlpaca}. Our approach is based on instruction-tuning on language-specific data augmented with a cross-lingual semantic alignment.
Hence, we set several baseline models (Section \ref{sec:It-LLMs-baseline}), which we augmented with our \textit{CrossAlpaca} approach (Section \ref{sec:It-LLMs-cross}). Finally, we performed a series of systematic evaluations (Section \ref{sec:benchmarks}) to observe the impact of the proposed intervention.

\subsection{Baseline Instruction-tuned LLMs}
\label{sec:It-LLMs-baseline}

The common denominator among the It-LLMs shown in Table \ref{tab:details_mono_ItLLMs} is the LLM backbone, LLaMA-7B \cite{touvron2023llama}. Starting from Instruction-following data from the original Alpaca \cite{alpaca} and its open-source non-English versions\footnote{open-source code is available on \url{https://github.com/tloen/alpaca-lora} \label{refnote}}, we reproduced \textit{x-Alpaca} for \textit{x} specific languages: Chinese (zh), Italian (it), Arabic (ar), Spanish (es), German (de) and the original English version (en).

\subsection{Cross-lingual Instruction-tuned LLMs}
\label{sec:It-LLMs-cross}

The \textit{CrossAlpacas} are instruction-tuned on Instruction-following and Translation-following demonstrations (\textit{CrossAlpacas demonstrations}). The first ones stem from the resources introduced in Section \ref{sec:It-LLMs-baseline}.  The second comes from \textit{news\_commentary} \cite{tiedemann-2012-parallel}.

Our approach generates a series of instruction-tuned versions of the data shown in Figure \ref{fig:our_proposal}. We have named the versions \textit{x-CrossAlpaca} where \textit{x} denotes: Chinese (zh), Arabic (ar), Italian (it), Spanish (es), and German (de).


\subsection{Experimental Setup}
\label{sec:exps_set}
In order to assess the performance of the \textit{CrossAlpaca}, we defined several benchmarks (Section \ref{sec:benchmarks}) on which we applied systematic tuning (Section \ref{sec:Models_Setup}) and evaluation (Section \ref{sec:Models_Evaluation}) pipelines.

\subsubsection{Benchmarks}
\label{sec:benchmarks}
To evaluate the performance of the It-LLMs and the 
impact of the translation-based semantic alignment approach, we used two cross-lingual (XQUAD \cite{Artetxe2019OnTC}, MLQA \cite{lewis-etal-2020-mlqa}) and two multi-task (MMLU \cite{hendrycks2021measuring} and BBH \cite{suzgun2022challenging}) benchmarks. 
XQUAD and MLQA focus on understanding questions and answers through translation into different languages. MMLU and BBH being multi-task benchmarks include subtasks related to Boolean expressions and QA on basic-level subjects (e.g., chemistry, physics).
However, we decided to introduce them to observe whether our approach degrades performance in these tasks. The first two datasets selected are appropriately constructed for multi-language testing, while the second two are available only in English. So we do a preliminary translation step as outlined below.

\paragraph{MultiLingual Question Answering (MLQA)} \cite{lewis-etal-2020-mlqa} evaluatates cross-lingual question answering performance. The benchmark comprises over 5K extractive QA instances in the SQuAD \cite{rajpurkar2016squad} format in several languages. MLQA is highly parallel, with QA instances aligned across four languages on average. Although comprising different languages, some languages are not represented, such as Italian. To conduct the experiments uniformly, we have translated the examples as also done in the forthcoming MMLU and BBH.

\paragraph{Cross-lingual Question Answering Dataset (XQUAD)} \cite{Artetxe2019OnTC} consists of a subset of 240 paragraphs and 1190 question-answer pairs from the development set of SQuAD v1.1 \cite{rajpurkar2016squad} with their manual translations into several languages. Consequently, the dataset is entirely parallel across 11 languages.

\paragraph{Massive Multitask Language Understanding (MMLU)} \cite{hendrycks2021measuring} measures knowledge of the world and problem-solving problems in multiple subjects with 57 subjects across STEM, humanities, social sciences, and other areas. The benchmark is native in English; however, we translated it into five additional languages\footnote{We performed translations using the Google translator API from English to Chinese (zh), Italian (it), Arabic (ar), Spanish (es), German (de). Resources will be made available along with the publication \label{languges}}.

\paragraph{BIG-Bench Hard (BBH)} \cite{suzgun2022challenging} is a subset of challenging tasks related to navigation, logical deduction, and fallacy detection. Here again, the benchmark is native English, and we have translated it into five languages\footref{languges}.

\subsubsection{Models Setup}
\label{sec:Models_Setup}
In order to align the results with the state-of-the-art models, we used the alpaca\_LoRA \cite{hu2021lora} code\footref{refnote}, adopting the same hyperparameters. 

We performed the fine-tuning with a single epoch and a batch-size of 128 examples, running our experiments on a workstation equipped with two Nvidia RTX A6000 with 48 GB of VRAM.

\subsubsection{Evaluation}
\label{sec:Models_Evaluation}
As an evaluation metric, we use accuracy. Hence, we estimate accuracy by measuring exact match values in the zero-shot setting. For each model, the parts of benchmarks related to the specific language are used (e.g., for zh-Alpaca and zh-CrossAlpaca, data from MLQA, XQUAD, MMLU, and BBH in Chinese are used).

\section{Results}
Improving non-English abilities in Instruction-tuned Large Language Models (It-LLMs) remains challenging. However, the \textit{x-CrossAlpacas} revealed improved results in cross-lingual Question Answering (QA) benchmarks and maintained logical-mathematical skills. From the results of Figure \ref{fig:performances} (further detailed in Table \ref{tab:tab_performances}), it is possible to observe the weaknesses emerging from the fine-tuning of the translated versions of Alpaca (Section \ref{sec:TranslatingIsNot}), the improvement obtained from the alignment phase is encouraging (Section \ref{sec:CrossAlpacaResults}) but it is not enough to outperform the English one. 
Therefore, we investigated the impact of demonstrations on downstream performance (Section \ref{sec:ablation}). 

The fine-grained analysis highlighted the importance of cross-lingual alignment data and the critical issues with non-English data. This opens the way for new hypotheses regarding the imbalance of pre-training languages and learning abilities via instruction-tuning.

\subsection{Translating the Alpaca is not the right way}
\label{sec:TranslatingIsNot}
The instruction-tuning on LLaMA, predominantly pre-trained in English, affects x-Alpaca. Figure \ref{fig:performances} and in terms of numbers, Table \ref{tab:tab_performances} show that in both MLQA and XQUAD, there is a gap of 55 and 53 average points between en-Alpaca and the x-Alpacas. This phenomenon is mitigated for MMLU and BBH, where we observed an average gap of 18 and 14 points. Instructing an LLM on Alpaca-style demonstrations translated into different languages is not always a good strategy. However, some x-Alpacas have performed better, such as zh-Alpaca and de-Alpaca. We hypothesize that this phenomenon is related to the scale of the pre-training data in the respective languages and, thus, the abilities of the LLaMA. In future developments, we plan to extend the study on other LLMs beyond LLaMA to observe whether the phenomenon is similar, milder, or more significant.

\begin{table}[h]
\small
\centering 
 \begin{tabular}{l||c|ccc}
\textbf{QA}  &  \textbf{en-} &  \textbf{avg-} & \textbf{avg-}  & \textbf{$\delta$}  \\
 \textbf{Task} &  \textbf{Alpaca} &  \textbf{Alpaca} & \textbf{CrossAlpaca}  &  \\

\hline
  &  &  & &  \\
MLQA  & \textit{0.89} &  0.34 & 0.64 & +0.30  \\
XQUAD  & \textit{0.97} & 0.31 & 0.65 & +0.30  \\
MMLU  & \textit{0.42} & 0.24 & 0.32 & +0.08  \\
BBH  & \textit{0.30}&  0.24 & 0.28 & +0.04  \\

\end{tabular}

\caption{Averages of the results on proposed benchmarks. The column $\delta$ indicates the difference between  avg-Alpacas and our avg-CrossAlpacas.}
\label{tab:evaluations_small}
\end{table}

\subsection{\textit{CrossAlpaca}: A cross-lingual solution}
\label{sec:CrossAlpacaResults}
Semantic alignment through Translation-following demonstrations during fine-tuning could have a valuable impact on the cross-lingual abilities of It-LLMs. The \textit{x-CrossAlpacas} outperformed the x-Alpacas by 30 average points on MLQA, 34 average points on XQUAD, 8 on MMLU, and 4 on BBH (see Table \ref{tab:evaluations_small} and more detailed in Table \ref{tab:tab_performances}). They also brought their performances closer to the sota obtained from the original Alpaca by 25 points on MLQA and 22 average points on XQUAD.  In MMLU and BBH, the gap became very close, with averages of 10 and 2 points (see Table \ref{tab:evaluations_small} or the line 'en-Alpaca vs avg-CrossAlpaca' in Table \ref{tab:tab_performances}).

Enriching with Translation-following demonstrations has outstanding influences on the cross-lingual abilities of the It-LLMs. However, even in this case, Chinese and German models (zh- and de-CrossAlpaca) outperformed Arabic by many points and, in some specific cases, Spanish and Italian as well.
This phenomenon, we hypothesize, is related to the divertsity in corpus representation within the pre-training data, as shown in \cite{yang2023bigtranslate}. Therefore, cross-lingual approaches do not have an incisive impact as in languages less present in the pre-training phases of the language model.

\begin{figure}[ht]
\centering
         \begin{minipage}{0.8\linewidth}
     \centering
     \includegraphics[width=\linewidth]{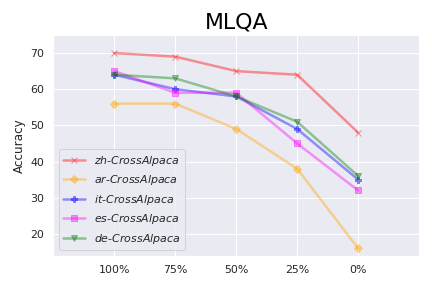}
   \end{minipage}
   \hfill
    \begin{minipage}{0.8\linewidth}
     \centering
  \includegraphics[width=\linewidth]{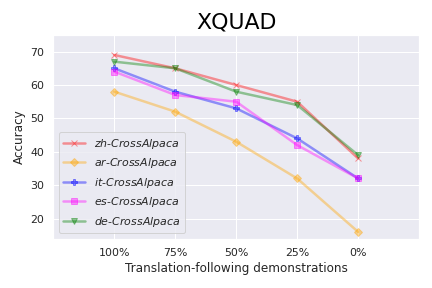}
   \end{minipage}
   \caption{Evaluation of proposed benchmarks of the demonstrations used for instruction-tuning our CrossAlpacas.} 
   \label{fig:Ablation_1}

\end{figure}

\subsection{Ablation Study}
\label{sec:ablation}
Our \textit{CrossAlpacas}, distinguished by the construction of the demonstrations pairs (Section \ref{sec:It-LLMs-cross}), 
achieves significant performance improvements and contributes to closing the gap between the original Alpaca (en-Alpaca) and a series of x-Alpacas in different languages. 
In order to show the impact of enrichment with cross-lingual demonstrations, we propose two different analyses. In the first analysis, we incrementally decrease the training data, in particular on the Translation-following demonstrations side (Section \ref{sec:Ablation_1}). In the second, still working on the Translation-following part (defined by half en-x and half x-en demonstrations), we analyze the impact of the demonstrations by splitting the experiments into en-x and x-en (Section \ref{sec:Ablation_2}).


\subsection{Translation-following demonstrations empower non-English abilities.}
\label{sec:Ablation_1}

Figure \ref{fig:Ablation_1} 
(for a more extensive analysis please refer to Figure \ref{fig:Ablation_1_complete} in the Appendix)
shows the non-English performance of the x-CrossAlpaca on the benchmarks under different scales of Translation-following demonstrations\footnote{The Traslation-following demonstrations are equally selected in a random way from the 10k en-x and x-en.}. The x-CrossAlpacas achieve higher accuracy when more demonstrations are used, revealing the benefits of cross-lingual alignment to improve non-English performance. However, Translation-following demonstrations are built in two directions: x-en and en-x (foreign-English and English-foreign). Although equally distributed, we are still determining whether there is an asymmetry between the two types of contributions. 

Finally, this result is less pronounced for the multi-task benchmarks (see Figure \ref{fig:Ablation_1_complete} MMLU and BBH). This could be related to the fact that in the benchmarks, there are logical-mathematical subtasks which are less language dependent (comparatively simpler syntax, smaller vocabularies), and hence, our cross-lingual approach does not influence the results for MLQA and XQUAD as much. Furthermore, it is possible to observe a trend in language groups, particularly Chinese-German, Italian-Spanish, and Arabic. This trend, which recurs frequently in the results of our experiments, will be further investigated in future work. 

\begin{figure}[h!]
\centering
         \begin{minipage}{0.80\linewidth}
     \centering
     \includegraphics[width=\linewidth]{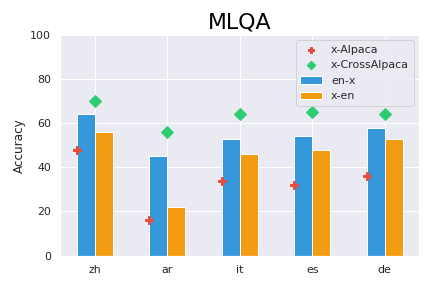}
   \end{minipage}
   \hfill
    \begin{minipage}{0.80\linewidth}
     \centering
  \includegraphics[width=\linewidth]{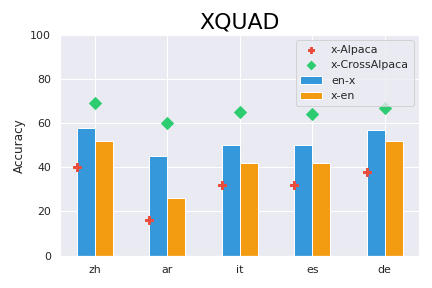}
   \end{minipage}
   \caption{Evaluation of proposed benchmarks using one-direction Translation-following demonstrations. For en-x for English-foreigner and x-en for foreign English.} 
   \label{fig:Ablation_2}

\end{figure}

\subsubsection{Demonstration direction matters non-English abilities}
\label{sec:Ablation_2}
Figure \ref{fig:Ablation_2} shows the evaluation conducted in varying directions of the translation-following demonstrations. In particular, demonstrations with English-foreign direction (en-x) better impact downstream models. Conversely, foreign-English (x-en) demonstrations perform better than baselines but underperform demonstrations in the opposite direction. 
However, as shown in Figure \ref{fig:Ablation_2} (and complete Figure \ref{fig:Ablation_2_complete}), the x-CrossAlpacas continue to outperform.
Nevertheless, the trend of translation-following demonstrations with one direction is interesting. Again, as was the case in the previous ablation study, the multi-task benchmarks (see MMLU and BBH in Figure \ref{fig:Ablation_2_complete}) do not seem to obtain significant influences, which reinforces the hypothesis that the models are greatly affected by cross-lingual capabilities in tasks where there is a strong presence of natural language.



\section{Future Works}
The cross-lingual abilities of Intruction-tuned Large Language Models (It-LLMs) seem to be supported by LLMs, such as in the case of Alpaca, the LLaMA backbone. However, it seems that low-impact demonstrations at the data level can enrich these abilities. We obtained valuable results from our experiments by proposing strategic demonstrations, i.e., Translation-following demonstrations.
These results were proposed by performing fine-tuning on LLaMA-7B as done by \citet{alpaca}.

In future work, we would like to continue to investigate by increasing the number of parameters in LLaMA and including additional backbone models. In addition, it might be interesting to evaluate the impact on low-resource languages. Hence, we would like to get to the underside of performances obtained in some experiments (see Section \ref{sec:Ablation_1}) by extending epistemic approaches \cite{ranaldi2023exploring} to It-LLMs.

In parallel, plans include analyzing the translation capabilities of general It-LLMs and those enhanced with translation tasks, including some specialized translation tasks among our evaluation benchmarks. 
Finally, we would like to investigate the learning abilities of the original Alpaca as the translation data change, proposing different probing experiments on (original) English data enhanced with translations.

\section{Conclusion}
In this paper, we proposed \textit{CrossAlpaca}, an approach to empowering the instruction-tuning of LLMs on non-English data. Specifically, we coupled Instruction-following (Alpaca-style) demonstrations with Translation-following demonstrations. Our method seeks to instruct the LLM to semantic alignment between English and non-English overperforms models instructed on non-English texts. In particular, thanks to our \textit{CrossAlpaca demonstrations}, the instructed models achieved significant performance improvements on four Question Answering benchmarks XQUAD, MLQA, MMLU, and BBH. In addition, we observe that semantic alignment strengthens with increasing Translation-following data; this demonstrates the de-facto abilities of It-LLMs to learn from instructions. Our approach and results contribute to improved research on the potential for producing more powerful LLMs for non-English languages.

\newpage



\bibliography{anthology,custom}

\begin{thebibliography}{47}
\expandafter\ifx\csname natexlab\endcsname\relax\def\natexlab#1{#1}\fi

\bibitem[{Abadji et~al.(2022)Abadji, Ortiz~Suarez, Romary, and
  Sagot}]{abadji-etal-2022-towards}
Julien Abadji, Pedro Ortiz~Suarez, Laurent Romary, and Beno{\^\i}t Sagot. 2022.
\newblock \href {https://aclanthology.org/2022.lrec-1.463} {Towards a cleaner
  document-oriented multilingual crawled corpus}.
\newblock In \emph{Proceedings of the Thirteenth Language Resources and
  Evaluation Conference}, pages 4344--4355, Marseille, France. European
  Language Resources Association.

\bibitem[{Artetxe et~al.(2019)Artetxe, Ruder, and Yogatama}]{Artetxe2019OnTC}
Mikel Artetxe, Sebastian Ruder, and Dani Yogatama. 2019.
\newblock \href {https://api.semanticscholar.org/CorpusID:204901567} {On the
  cross-lingual transferability of monolingual representations}.
\newblock In \emph{Annual Meeting of the Association for Computational
  Linguistics}.

\bibitem[{Aulamo and Tiedemann(2019)}]{aulamo-tiedemann-2019-opus}
Mikko Aulamo and J{\"o}rg Tiedemann. 2019.
\newblock \href {https://aclanthology.org/W19-6146} {The {OPUS} resource
  repository: An open package for creating parallel corpora and machine
  translation services}.
\newblock In \emph{Proceedings of the 22nd Nordic Conference on Computational
  Linguistics}, pages 389--394, Turku, Finland. Link{\"o}ping University
  Electronic Press.

\bibitem[{Bang et~al.(2023)Bang, Cahyawijaya, Lee, Dai, Su, Wilie, Lovenia, Ji,
  Yu, Chung, Do, Xu, and Fung}]{bang2023multitask}
Yejin Bang, Samuel Cahyawijaya, Nayeon Lee, Wenliang Dai, Dan Su, Bryan Wilie,
  Holy Lovenia, Ziwei Ji, Tiezheng Yu, Willy Chung, Quyet~V. Do, Yan Xu, and
  Pascale Fung. 2023.
\newblock \href {http://arxiv.org/abs/2302.04023} {A multitask, multilingual,
  multimodal evaluation of chatgpt on reasoning, hallucination, and
  interactivity}.

\bibitem[{Blevins and Zettlemoyer(2022)}]{Blevins2022LanguageCH}
Terra Blevins and Luke Zettlemoyer. 2022.
\newblock \href {https://api.semanticscholar.org/CorpusID:252780005} {Language
  contamination helps explains the cross-lingual capabilities of english
  pretrained models}.
\newblock In \emph{Conference on Empirical Methods in Natural Language
  Processing}.

\bibitem[{Brown et~al.(2020)Brown, Mann, Ryder, Subbiah, Kaplan, Dhariwal,
  Neelakantan, Shyam, Sastry, Askell, Agarwal, Herbert-Voss, Krueger, Henighan,
  Child, Ramesh, Ziegler, Wu, Winter, Hesse, Chen, Sigler, Litwin, Gray, Chess,
  Clark, Berner, McCandlish, Radford, Sutskever, and
  Amodei}]{brown2020language}
Tom~B. Brown, Benjamin Mann, Nick Ryder, Melanie Subbiah, Jared Kaplan,
  Prafulla Dhariwal, Arvind Neelakantan, Pranav Shyam, Girish Sastry, Amanda
  Askell, Sandhini Agarwal, Ariel Herbert-Voss, Gretchen Krueger, Tom Henighan,
  Rewon Child, Aditya Ramesh, Daniel~M. Ziegler, Jeffrey Wu, Clemens Winter,
  Christopher Hesse, Mark Chen, Eric Sigler, Mateusz Litwin, Scott Gray,
  Benjamin Chess, Jack Clark, Christopher Berner, Sam McCandlish, Alec Radford,
  Ilya Sutskever, and Dario Amodei. 2020.
\newblock \href {http://arxiv.org/abs/2005.14165} {Language models are few-shot
  learners}.

\bibitem[{Chen et~al.(2023)Chen, Wu, and Chen}]{traditional-chinese-alpaca}
Wei-Lin Chen, Cheng-Kuang Wu, and Hsin-Hsi Chen. 2023.
\newblock Traditional-chinese alpaca: Models and datasets.
\newblock \url{https://github.com/ntunlplab/traditional-chinese-alpaca}.

\bibitem[{Cui et~al.(2023)Cui, Yang, and Yao}]{cui2023efficient}
Yiming Cui, Ziqing Yang, and Xin Yao. 2023.
\newblock \href {http://arxiv.org/abs/2304.08177} {Efficient and effective text
  encoding for chinese llama and alpaca}.

\bibitem[{Dodge et~al.(2021)Dodge, Sap, Marasović, Agnew, Ilharco, Groeneveld,
  Mitchell, and Gardner}]{dodge2021documenting}
Jesse Dodge, Maarten Sap, Ana Marasović, William Agnew, Gabriel Ilharco, Dirk
  Groeneveld, Margaret Mitchell, and Matt Gardner. 2021.
\newblock \href {http://arxiv.org/abs/2104.08758} {Documenting large webtext
  corpora: A case study on the colossal clean crawled corpus}.

\bibitem[{Gao et~al.(2020)Gao, Biderman, Black, Golding, Hoppe, Foster, Phang,
  He, Thite, Nabeshima, Presser, and Leahy}]{gao2020pile}
Leo Gao, Stella Biderman, Sid Black, Laurence Golding, Travis Hoppe, Charles
  Foster, Jason Phang, Horace He, Anish Thite, Noa Nabeshima, Shawn Presser,
  and Connor Leahy. 2020.
\newblock \href {http://arxiv.org/abs/2101.00027} {The pile: An 800gb dataset
  of diverse text for language modeling}.

\bibitem[{Hendrycks et~al.(2021)Hendrycks, Burns, Basart, Zou, Mazeika, Song,
  and Steinhardt}]{hendrycks2021measuring}
Dan Hendrycks, Collin Burns, Steven Basart, Andy Zou, Mantas Mazeika, Dawn
  Song, and Jacob Steinhardt. 2021.
\newblock \href {http://arxiv.org/abs/2009.03300} {Measuring massive multitask
  language understanding}.

\bibitem[{Hu et~al.(2021{\natexlab{a}})Hu, Shen, Wallis, Allen-Zhu, Li, Wang,
  Wang, and Chen}]{hu2022lora}
Edward~J. Hu, Yelong Shen, Phillip Wallis, Zeyuan Allen-Zhu, Yuanzhi Li, Shean
  Wang, Lu~Wang, and Weizhu Chen. 2021{\natexlab{a}}.
\newblock \href {http://arxiv.org/abs/2106.09685} {Lora: Low-rank adaptation of
  large language models}.

\bibitem[{Hu et~al.(2021{\natexlab{b}})Hu, Shen, Wallis, Allen-Zhu, Li, Wang,
  Wang, and Chen}]{hu2021lora}
Edward~J. Hu, Yelong Shen, Phillip Wallis, Zeyuan Allen-Zhu, Yuanzhi Li, Shean
  Wang, Lu~Wang, and Weizhu Chen. 2021{\natexlab{b}}.
\newblock \href {http://arxiv.org/abs/2106.09685} {Lora: Low-rank adaptation of
  large language models}.

\bibitem[{Huang et~al.(2023)Huang, Tang, Zhang, Zhao, Song, Xia, and
  Wei}]{huang2023languages}
Haoyang Huang, Tianyi Tang, Dongdong Zhang, Wayne~Xin Zhao, Ting Song, Yan Xia,
  and Furu Wei. 2023.
\newblock \href {http://arxiv.org/abs/2305.07004} {Not all languages are
  created equal in llms: Improving multilingual capability by
  cross-lingual-thought prompting}.

\bibitem[{Imani et~al.(2023)Imani, Lin, Kargaran, Severini, Sabet, Kassner, Ma,
  Schmid, Martins, Yvon, and Schütze}]{imani2023glot500}
Ayyoob Imani, Peiqin Lin, Amir~Hossein Kargaran, Silvia Severini, Masoud~Jalili
  Sabet, Nora Kassner, Chunlan Ma, Helmut Schmid, André F.~T. Martins,
  François Yvon, and Hinrich Schütze. 2023.
\newblock \href {http://arxiv.org/abs/2305.12182} {Glot500: Scaling
  multilingual corpora and language models to 500 languages}.

\bibitem[{{Kohaku-Blueleaf}(2023)}]{Guanaco-lora}
{Kohaku-Blueleaf}. 2023.
\newblock {Guanaco-lora: LoRA for trainin Multilingual Instruction-following LM
  based on LLaMA}.
\newblock {https://huggingface.co/plncmm/guanaco-lora-7b }.

\bibitem[{Lahiri(2014)}]{lahiri:2014:SRW}
Shibamouli Lahiri. 2014.
\newblock \href {http://www.aclweb.org/anthology/E14-3011} {{Complexity of Word
  Collocation Networks: A Preliminary Structural Analysis}}.
\newblock In \emph{Proceedings of the Student Research Workshop at the 14th
  Conference of the European Chapter of the Association for Computational
  Linguistics}, pages 96--105, Gothenburg, Sweden. Association for
  Computational Linguistics.

\bibitem[{Le et~al.(2021)Le, Pino, Wang, Gu, Schwab, and
  Besacier}]{le2021lightweight}
Hang Le, Juan~Miguel Pino, Changhan Wang, Jiatao Gu, Didier Schwab, and Laurent
  Besacier. 2021.
\newblock Lightweight adapter tuning for multilingual speech translation.
\newblock In \emph{Proceedings of the 59th Annual Meeting of the Association
  for Computational Linguistics (ACL)}, pages 817--824. Association for
  Computational Linguistics.

\bibitem[{Lewis et~al.(2020)Lewis, Oguz, Rinott, Riedel, and
  Schwenk}]{lewis-etal-2020-mlqa}
Patrick Lewis, Barlas Oguz, Ruty Rinott, Sebastian Riedel, and Holger Schwenk.
  2020.
\newblock \href {https://doi.org/10.18653/v1/2020.acl-main.653} {{MLQA}:
  Evaluating cross-lingual extractive question answering}.
\newblock In \emph{Proceedings of the 58th Annual Meeting of the Association
  for Computational Linguistics}, pages 7315--7330, Online. Association for
  Computational Linguistics.

\bibitem[{Li et~al.(2023)Li, Zhou, Huang, Cheng, and Chen}]{li2023eliciting}
Jiahuan Li, Hao Zhou, Shujian Huang, Shanbo Cheng, and Jiajun Chen. 2023.
\newblock \href {http://arxiv.org/abs/2305.15083} {Eliciting the translation
  ability of large language models via multilingual finetuning with translation
  instructions}.

\bibitem[{Li and Liang(2021)}]{li-liang-2021-prefix}
Xiang~Lisa Li and Percy Liang. 2021.
\newblock \href {https://doi.org/10.18653/v1/2021.acl-long.353} {Prefix-tuning:
  Optimizing continuous prompts for generation}.
\newblock In \emph{Proceedings of the 59th Annual Meeting of the Association
  for Computational Linguistics and the 11th International Joint Conference on
  Natural Language Processing (Volume 1: Long Papers)}, pages 4582--4597,
  Online. Association for Computational Linguistics.

\bibitem[{Lin et~al.(2021)Lin, Pan, Wang, Qiu, Feng, Zhou, and
  Li}]{lin2021pretraining}
Zehui Lin, Xiao Pan, Mingxuan Wang, Xipeng Qiu, Jiangtao Feng, Hao Zhou, and
  Lei Li. 2021.
\newblock \href {http://arxiv.org/abs/2010.03142} {Pre-training multilingual
  neural machine translation by leveraging alignment information}.

\bibitem[{Liu et~al.(2022)Liu, Ji, Fu, Tam, Du, Yang, and
  Tang}]{liu-etal-2022-p}
Xiao Liu, Kaixuan Ji, Yicheng Fu, Weng Tam, Zhengxiao Du, Zhilin Yang, and Jie
  Tang. 2022.
\newblock \href {https://doi.org/10.18653/v1/2022.acl-short.8} {{P}-tuning:
  Prompt tuning can be comparable to fine-tuning across scales and tasks}.
\newblock In \emph{Proceedings of the 60th Annual Meeting of the Association
  for Computational Linguistics (Volume 2: Short Papers)}, pages 61--68,
  Dublin, Ireland. Association for Computational Linguistics.

\bibitem[{Moslem et~al.(2023)Moslem, Haque, Kelleher, and
  Way}]{moslem2023adaptive}
Yasmin Moslem, Rejwanul Haque, John~D. Kelleher, and Andy Way. 2023.
\newblock \href {http://arxiv.org/abs/2301.13294} {Adaptive machine translation
  with large language models}.

\bibitem[{Ouyang et~al.(2022)Ouyang, Wu, Jiang, Almeida, Wainwright, Mishkin,
  Zhang, Agarwal, Slama, Ray, Schulman, Hilton, Kelton, Miller, Simens, Askell,
  Welinder, Christiano, Leike, and Lowe}]{ouyang2022training}
Long Ouyang, Jeff Wu, Xu~Jiang, Diogo Almeida, Carroll~L. Wainwright, Pamela
  Mishkin, Chong Zhang, Sandhini Agarwal, Katarina Slama, Alex Ray, John
  Schulman, Jacob Hilton, Fraser Kelton, Luke Miller, Maddie Simens, Amanda
  Askell, Peter Welinder, Paul Christiano, Jan Leike, and Ryan Lowe. 2022.
\newblock \href {http://arxiv.org/abs/2203.02155} {Training language models to
  follow instructions with human feedback}.

\bibitem[{Penedo et~al.(2023)Penedo, Malartic, Hesslow, Cojocaru, Cappelli,
  Alobeidli, Pannier, Almazrouei, and Launay}]{penedo2023refinedweb}
Guilherme Penedo, Quentin Malartic, Daniel Hesslow, Ruxandra Cojocaru,
  Alessandro Cappelli, Hamza Alobeidli, Baptiste Pannier, Ebtesam Almazrouei,
  and Julien Launay. 2023.
\newblock \href {http://arxiv.org/abs/2306.01116} {The refinedweb dataset for
  falcon llm: Outperforming curated corpora with web data, and web data only}.

\bibitem[{Rajpurkar et~al.(2016)Rajpurkar, Zhang, Lopyrev, and
  Liang}]{rajpurkar2016squad}
Pranav Rajpurkar, Jian Zhang, Konstantin Lopyrev, and Percy Liang. 2016.
\newblock \href {http://arxiv.org/abs/1606.05250} {Squad: 100,000+ questions
  for machine comprehension of text}.

\bibitem[{Ranaldi et~al.(2023)Ranaldi, Ruzzetti, Logozzo, Mastromattei,
  Ranaldi, and Zanzotto}]{ranaldi2023exploring}
Federico Ranaldi, Elena~Sofia Ruzzetti, Felicia Logozzo, Michele Mastromattei,
  Leonardo Ranaldi, and Fabio~Massimo Zanzotto. 2023.
\newblock \href {http://arxiv.org/abs/2305.02215} {Exploring linguistic
  properties of monolingual berts with typological classification among
  languages}.

\bibitem[{Santilli and Rodolà(2023)}]{santilli2023camoscio}
Andrea Santilli and Emanuele Rodolà. 2023.
\newblock \href {http://arxiv.org/abs/2307.16456} {Camoscio: an italian
  instruction-tuned llama}.

\bibitem[{Shoeybi et~al.(2019)Shoeybi, Patwary, Puri, LeGresley, Casper, and
  Catanzaro}]{Shoeybi2019MegatronLMTM}
Mohammad Shoeybi, Mostofa~Ali Patwary, Raul Puri, Patrick LeGresley, Jared
  Casper, and Bryan Catanzaro. 2019.
\newblock Megatron-{LM}: Training multi-billion parameter language models using
  model parallelism.
\newblock \emph{ArXiv}, abs/1909.08053.

\bibitem[{Suzgun et~al.(2022)Suzgun, Scales, Sch{\"a}rli, Gehrmann, Tay, Chung,
  Chowdhery, Le, Chi, Zhou, , and Wei}]{suzgun2022challenging}
Mirac Suzgun, Nathan Scales, Nathanael Sch{\"a}rli, Sebastian Gehrmann, Yi~Tay,
  Hyung~Won Chung, Aakanksha Chowdhery, Quoc~V Le, Ed~H Chi, Denny Zhou, , and
  Jason Wei. 2022.
\newblock Challenging big-bench tasks and whether chain-of-thought can solve
  them.
\newblock \emph{arXiv preprint arXiv:2210.09261}.

\bibitem[{Tanti et~al.(2021)Tanti, van~der Plas, Borg, and
  Gatt}]{tanti2021languagespecificity}
Marc Tanti, Lonneke van~der Plas, Claudia Borg, and Albert Gatt. 2021.
\newblock \href {http://arxiv.org/abs/2109.06935} {On the language-specificity
  of multilingual bert and the impact of fine-tuning}.

\bibitem[{Taori et~al.(2023)Taori, Gulrajani, Zhang, Dubois, Li, Guestrin,
  Liang, and Hashimoto}]{alpaca}
Rohan Taori, Ishaan Gulrajani, Tianyi Zhang, Yann Dubois, Xuechen Li, Carlos
  Guestrin, Percy Liang, and Tatsunori~B. Hashimoto. 2023.
\newblock Stanford alpaca: An instruction-following llama model.
\newblock \url{https://github.com/tatsu-lab/stanford_alpaca}.

\bibitem[{Tenney et~al.(2019)Tenney, Das, and Pavlick}]{tenney-etal-2019-bert}
Ian Tenney, Dipanjan Das, and Ellie Pavlick. 2019.
\newblock \href {https://doi.org/10.18653/v1/P19-1452} {{BERT} rediscovers the
  classical {NLP} pipeline}.
\newblock In \emph{Proceedings of the 57th Annual Meeting of the Association
  for Computational Linguistics}, pages 4593--4601, Florence, Italy.
  Association for Computational Linguistics.

\bibitem[{Thissen(2023)}]{de_alpaca}
Martin Thissen. 2023.
\newblock Fine-tune alpaca for any language.
\newblock \url{https://github.com/thisserand/alpaca-lora-finetune-language}.

\bibitem[{Tiedemann(2012)}]{tiedemann-2012-parallel}
J{\"o}rg Tiedemann. 2012.
\newblock \href
  {http://www.lrec-conf.org/proceedings/lrec2012/pdf/463_Paper.pdf} {Parallel
  data, tools and interfaces in {OPUS}}.
\newblock In \emph{Proceedings of the Eighth International Conference on
  Language Resources and Evaluation ({LREC}'12)}, pages 2214--2218, Istanbul,
  Turkey. European Language Resources Association (ELRA).

\bibitem[{Touvron et~al.(2023)Touvron, Lavril, Izacard, Martinet, Lachaux,
  Lacroix, Rozière, Goyal, Hambro, Azhar, Rodriguez, Joulin, Grave, and
  Lample}]{touvron2023llama}
Hugo Touvron, Thibaut Lavril, Gautier Izacard, Xavier Martinet, Marie-Anne
  Lachaux, Timothée Lacroix, Baptiste Rozière, Naman Goyal, Eric Hambro,
  Faisal Azhar, Aurelien Rodriguez, Armand Joulin, Edouard Grave, and Guillaume
  Lample. 2023.
\newblock \href {http://arxiv.org/abs/2302.13971} {Llama: Open and efficient
  foundation language models}.

\bibitem[{Wang et~al.(2023)Wang, Kordi, Mishra, Liu, Smith, Khashabi, and
  Hajishirzi}]{wang2023selfinstruct}
Yizhong Wang, Yeganeh Kordi, Swaroop Mishra, Alisa Liu, Noah~A. Smith, Daniel
  Khashabi, and Hannaneh Hajishirzi. 2023.
\newblock \href {http://arxiv.org/abs/2212.10560} {Self-instruct: Aligning
  language models with self-generated instructions}.

\bibitem[{Wei et~al.(2022)Wei, Bosma, Zhao, Guu, Yu, Lester, Du, Dai, and
  Le}]{wei2022finetuned}
Jason Wei, Maarten Bosma, Vincent~Y. Zhao, Kelvin Guu, Adams~Wei Yu, Brian
  Lester, Nan Du, Andrew~M. Dai, and Quoc~V. Le. 2022.
\newblock \href {http://arxiv.org/abs/2109.01652} {Finetuned language models
  are zero-shot learners}.

\bibitem[{Xu et~al.(2023)Xu, Guo, Duan, and McAuley}]{xu2023baize}
Canwen Xu, Daya Guo, Nan Duan, and Julian McAuley. 2023.
\newblock Baize: An open-source chat model with parameter-efficient tuning on
  self-chat data.
\newblock \emph{arXiv preprint arXiv:2304.01196}.

\bibitem[{Yang et~al.(2023)Yang, Li, Zhang, and Zong}]{yang2023bigtranslate}
Wen Yang, Chong Li, Jiajun Zhang, and Chengqing Zong. 2023.
\newblock \href {http://arxiv.org/abs/2305.18098} {Bigtranslate: Augmenting
  large language models with multilingual translation capability over 100
  languages}.

\bibitem[{{Yasbok}()}]{yasbok-alpaca-instruction-fine-tune-arabic}
{Yasbok}.
\newblock {Alpaca Instruction Fine-Tuning for Arabic}.
\newblock {https://huggingface.co/Yasbok }.

\bibitem[{Zeng et~al.(2023)Zeng, Meng, Yin, and Zhou}]{zeng2023tim}
Jiali Zeng, Fandong Meng, Yongjing Yin, and Jie Zhou. 2023.
\newblock \href {http://arxiv.org/abs/2307.04408} {Tim: Teaching large language
  models to translate with comparison}.

\bibitem[{Zhang et~al.(2023)Zhang, Fang, Zhang, Ma, Zhou, Huang, Bu, Gui, Chen,
  Chen, and Feng}]{zhang2023bayling}
Shaolei Zhang, Qingkai Fang, Zhuocheng Zhang, Zhengrui Ma, Yan Zhou, Langlin
  Huang, Mengyu Bu, Shangtong Gui, Yunji Chen, Xilin Chen, and Yang Feng. 2023.
\newblock \href {http://arxiv.org/abs/2306.10968} {Bayling: Bridging
  cross-lingual alignment and instruction following through interactive
  translation for large language models}.

\bibitem[{Zhu et~al.(2023{\natexlab{a}})Zhu, Liu, Dong, Xu, Huang, Kong, Chen,
  and Li}]{zhu2023multilingual}
Wenhao Zhu, Hongyi Liu, Qingxiu Dong, Jingjing Xu, Shujian Huang, Lingpeng
  Kong, Jiajun Chen, and Lei Li. 2023{\natexlab{a}}.
\newblock \href {http://arxiv.org/abs/2304.04675} {Multilingual machine
  translation with large language models: Empirical results and analysis}.

\bibitem[{Zhu et~al.(2023{\natexlab{b}})Zhu, Lv, Dong, Yuan, Xu, Huang, Kong,
  Chen, and Li}]{zhu2023extrapolating}
Wenhao Zhu, Yunzhe Lv, Qingxiu Dong, Fei Yuan, Jingjing Xu, Shujian Huang,
  Lingpeng Kong, Jiajun Chen, and Lei Li. 2023{\natexlab{b}}.
\newblock \href {http://arxiv.org/abs/2308.04948} {Extrapolating large language
  models to non-english by aligning languages}.

\bibitem[{Zhu et~al.(2015)Zhu, Kiros, Zemel, Salakhutdinov, Urtasun, Torralba,
  and Fidler}]{Zhu_2015_ICCV}
Yukun Zhu, Ryan Kiros, Rich Zemel, Ruslan Salakhutdinov, Raquel Urtasun,
  Antonio Torralba, and Sanja Fidler. 2015.
\newblock Aligning books and movies: Towards story-like visual explanations by
  watching movies and reading books.
\newblock In \emph{The IEEE International Conference on Computer Vision
  (ICCV)}.

\end{thebibliography}
\bibliographystyle{acl_natbib} 

\newpage
\appendix

\begin{table*}[t]
\section{Appendix}

\centering 

\begin{tabular}{l||l|c|c|c|c}
\hline

& \textbf{Model} & \textbf{MLQA} & \textbf{XQUAD} & \textbf{MMLU} & \textbf{BBH}  \\  

\hline
\hline

\multirow{7}{*}{\begin{sideways}\textbf{\textit{Alpaca}}\end{sideways}}
& \textit{en-Alpaca} & \textit{0.89} & \textit{0.87} & \textit{0.42} & \textit{0.30}   \\
\hline
& zh-Alpaca & \textbf{0.48} & 0.38 & \textbf{0.26} & \textbf{0.25}   \\
& ar-Alpaca & 0.17 & 0.16 & 0.18 & 0.20  \\
& it-Alpaca & 0.35 & 0.32 & 0.21 & 0.25   \\
& es-Alpaca & 0.32 & 0.33 & 0.19 & 0.24   \\
& de-Alpaca & 0.36 & \textbf{0.39} & 0.24 & 0.25   \\
\hline
& \textbf{\textit{avg}-Alpaca} & 0.34 & 0.31 & 0.24 & 0.24   \\

\hline
\multirow{5}{*}{\begin{sideways}\textbf{\textit{CrossAlpaca}}\end{sideways}}

& zh-CrossAlpaca & \textbf{0.70} & \textbf{0.69} & \textbf{0.36} & 0.28  \\
& ar-CrossAlpaca & 0.56 & 0.60 & 0.25 & 0.25   \\
& it-CrossAlpaca & 0.64 & 0.65 & 0.28 & 0.27   \\
& es-CrossAlpaca & 0.65 & 0.64 & 0.28 & 0.28   \\
& de-CrossAlpaca & 0.64 & 0.67 & 0.32  & \textbf{0.29}   \\
\hline
& \textbf{\textit{avg}-CrossAlpaca} & 0.64 & 0.65 & 0.32 & 0.28   \\

\hline

&  &  &  &  &    \\

\multirow{3}{*}{\begin{sideways}\end{sideways}}
& \textbf{en-Alpaca vs \textit{avg}-Alpaca} & 0.34\textit{\textbf{(-0.55)}} & 0.31\textit{\textbf{(-0.56)}} & 0.24\textit{\textbf{(-0.18)}} & 0.24\textit{\textbf{(-0.06)}}   \\
& \textbf{en-Alpaca vs \textit{avg}-CrossAlpaca} & 0.64\textit{\textbf{(-0.25)}} & 0.65\textit{\textbf{(-0.22)}} & 0.32\textit{\textbf{(-0.10)}} & 0.28\textit{\textbf{(-0.20)}}   \\
\hline
& \textbf{\textit{avg}-CrossAlpaca vs \textit{avg}-Alpaca} & \textit{\textbf{(+0.30)}} & \textit{\textbf{(+0.34)}} & \textit{\textbf{(+0.08)}} & \textit{\textbf{(+0.04)}}   \\
\hline

\hline 

\hline
\end{tabular}

 \caption{Evaluation results on proposed benchmarks. The \textit{\textbf{avg-}} lines are the averages of x-Alpacas and x-CorssAlpacas. The last three lines indicate the comparisons between en-Alpaca. The last line indicates the comparisons between avg-Alpacas and avg-CrossAlpacas.}
 \label{tab:tab_performances}

\end{table*}

\begin{table*}[t]
\section{Appendix}
\label{sec:Appendix3}
\centering 
 \begin{tabular}{l|cccccc}
\textbf{Language}  & \textbf{Alpaca} & \textbf{MLQA} & \textbf{XQUAD} & \textbf{MMLU} & \textbf{BBH} & \textbf{News\_commentary} \\

\hline
\textbf{Arabic} & \textit{x} & \textit{x} & \textit{x} & \textit{-} & \textit{-} & \textit{x} \\
\textbf{Chinese} & \textit{x} & \textit{x} & \textit{x} & \textit{-} & \textit{-} & \textit{x} \\
\textbf{English}  & \textit{x} & \textit{x} & \textit{x} & \textit{x} & \textit{x} & \textit{x}  \\
\textbf{German} & \textit{x} & \textit{x} & \textit{x} & \textit{-} & \textit{-} & \textit{x} \\
Greek & \textit{-} & \textit{x} & \textit{x} & \textit{-} & \textit{-} & \textit{x} \\
Hindi & \textit{x} & \textit{x} & \textit{x} & \textit{-} & \textit{-} & \textit{x} \\
\textbf{Italian} & \textit{x} & \textit{-} & \textit{-} & \textit{x} & \textit{x} & \textit{x}  \\
Russian & \textit{x} & \textit{x} & \textit{x} & \textit{-} & \textit{-} & \textit{x} \\
\textbf{Spanish} & \textit{x} & \textit{x} & \textit{x} & \textit{-} & \textit{-} & \textit{x} \\
Turkish & \textit{x} & \textit{-} & \textit{x} & \textit{-} & \textit{-} & \textit{-} \\
Vietnamese & \textit{-} & \textit{x} & \textit{x} & \textit{-} & \textit{-} & \textit{-} \\

 \end{tabular}

 \caption{List of available state-of-the-art resources.}
 \label{tab:details_data_available}
\end{table*}

\begin{figure*}[t]
\section{Appendix}

\centering
         \begin{minipage}{0.45\linewidth}
     \centering
     \includegraphics[width=\linewidth]{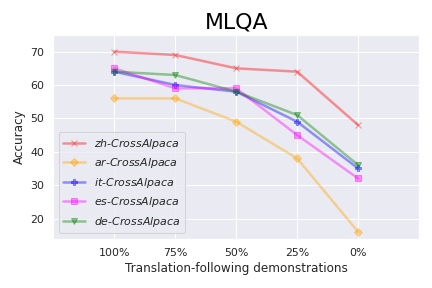}
   \end{minipage}
    \begin{minipage}{0.45\linewidth}
     \centering
  \includegraphics[width=\linewidth]{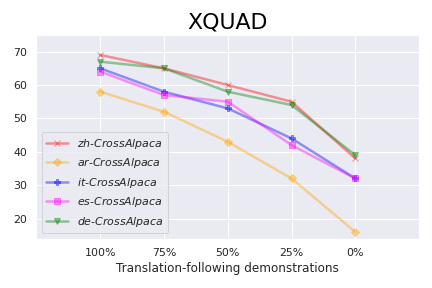}
   \end{minipage}
         \begin{minipage}{0.45\linewidth}
     \centering
     \includegraphics[width=\linewidth]{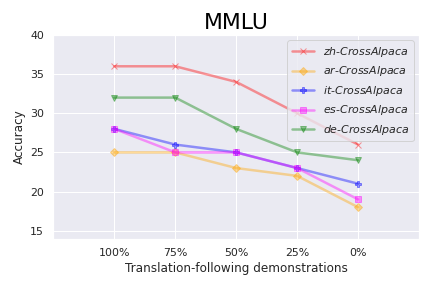}
   \end{minipage}
            \begin{minipage}{0.45\linewidth}
     \centering
     \includegraphics[width=\linewidth]{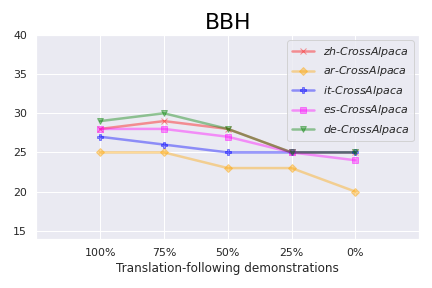}
   \end{minipage}
   \caption{Evaluation of all proposed benchmarks of the demonstrations used for instruction-tuning our CrossAlpacas.} 
   \label{fig:Ablation_1_complete}

\end{figure*}

\begin{figure*}[t]
\section{Appendix}

\centering
         \begin{minipage}{0.45\linewidth}
     \centering
     \includegraphics[width=\linewidth]{img/MLQA_en_x.png}
   \end{minipage}
    \begin{minipage}{0.45\linewidth}
     \centering
  \includegraphics[width=\linewidth]{img/XQUAD_en_x.png}
   \end{minipage}
         \begin{minipage}{0.45\linewidth}
     \centering
     \includegraphics[width=\linewidth]{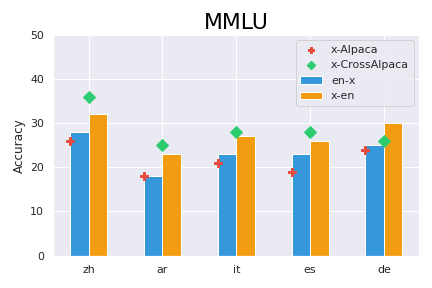}
   \end{minipage}
            \begin{minipage}{0.45\linewidth}
     \centering
     \includegraphics[width=\linewidth]{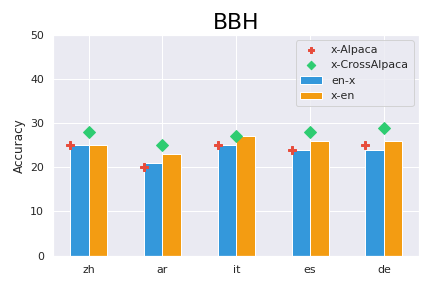}
   \end{minipage}
   \caption{Evaluation of all proposed benchmarks using one-direction Translation-following demonstrations. For en-x for English-foreigner and x-en for foreign English. With the red cross, we indicate the results of the x-Alpaca standards, and with the green diamond, the results of our x-CrossAlpaca.} 
   \label{fig:Ablation_2_complete}

\end{figure*}

\end{document}